\documentclass[lettersize,journal]{IEEEtran}
\usepackage{amsmath,amsfonts}
\usepackage{array}
\usepackage[caption=false,font=normalsize,labelfont=sf,textfont=sf]{subfig}
\usepackage{textcomp}
\usepackage{stfloats}
\usepackage{url}
\usepackage{verbatim}
\usepackage{graphicx}
\usepackage{cite}
\usepackage[ruled,vlined]{algorithm2e}
\usepackage{setspace}
\usepackage{booktabs}
\usepackage{adjustbox}
\usepackage[table]{xcolor}
\usepackage{multirow}
\definecolor{groupgray}{RGB}{247,247,248}
\definecolor{headergray}{RGB}{235,236,238}
\definecolor{oursbg}{RGB}{235,246,252}
\definecolor{oursblue}{RGB}{32,82,132}
\definecolor{deltared}{RGB}{235,35,35}
\usepackage{amsthm}
\newtheorem{assumption}{Assumption}
\newtheorem{lemma}{Lemma}
\newtheorem{theorem}{Theorem}
\usepackage{amssymb}
\begin{document}

\title{P-FedSkill: A Personalized Federated Skill Learning Framework under Task Heterogeneity}

\author{Zehang Lin, Jiayi Ji~\IEEEmembership{Senior Member,~IEEE}, and Rongrong Ji,~\IEEEmembership{Senior Member,~IEEE}%
\thanks{Z. Lin, J. Ji, and R. Ji are with the Key Laboratory of Multimedia
Trusted Perception and Efficient Computing, Ministry of Education of China,
School of Informatics, Xiamen University, Xiamen 361005, China
(e-mail: linzehang@stu.xmu.edu.cn; jjyxmu@gmail.com; rrji@xmu.edu.cn).}%
}


\maketitle

\begin{abstract}
Skill learning has become an important way to improve agent performance. However, agent trajectories often contain sensitive information and cannot be shared across clients, hindering collaborative skill learning.
To tackle this issue, federated skill learning (FSL) has emerged as a solution by distilling trajectories into compact skill representations and uploading them to a central server for sharing.
However, existing FSL methods typically share skills based on skill similarity without considering their contributions to heterogeneous tasks across clients, thereby degrading agent performance.
To address this challenge, we propose P-FedSkill, a personalized FSL framework that comprises three key components: local skill extraction (LSE), personalized skill selection (PSS), and task-aware skill routing (TASR). 
Specifically, LSE first extracts skill patches (i.e., task applicability conditions and execution strategies) and task execution outcomes from trajectories.
PSS then uses these outcomes to evaluate each skill patch's performance on client tasks and incorporates its task applicability condition to construct a personalized skill set for each client.
Finally, TASR adaptively selects a task-relevant subset of skill patches based on their applicability conditions to guide agent execution.
We further provide a theoretical guarantee for personalized skill selection in P-FedSkill, and extensive experiments also demonstrate its effectiveness.
\end{abstract}


\section{Introduction}
\label{sec:intro}
\IEEEPARstart{W}{ith} the proliferation of intelligent agents, massive volumes of interaction trajectories are being generated on edge devices~\cite{wang2025distrl,lv2026m2miner}. According to the International Data Corporation (IDC), more than 1 billion AI agents are expected to be deployed globally by 2029, performing approximately 217 billion actions daily~\cite{villars2025agent}. This rapidly growing volume of interaction data is accelerating advances in agent technologies, leading to progress across various domains such as intelligent manufacturing~\cite{chen2025opc}, cybersecurity~\cite{deng2024pentestgpt}, and embodied intelligence~\cite{hong2024multiply}.

Recently, skill learning has become an important approach to improving agent performance by extracting reusable task solutions from interaction trajectories~\cite{yu2026polyskill}. These solutions can be further used to guide agents in completing new tasks, thereby improving execution efficiency~\cite{wang2025awm}. However, trajectories from individual clients typically span only a narrow range of tasks, limiting the solutions available for agent execution~\cite{jiang2023fedskill}. A straightforward idea is to collect trajectories from multiple clients to learn a broader set of skills. However, such trajectories often contain sensitive information and cannot be directly shared across clients due to privacy concerns, thereby hindering collaborative skill learning~\cite{zhang2026searching}.

\begin{figure}[t]
    \centering
    \includegraphics[width=0.95\linewidth]{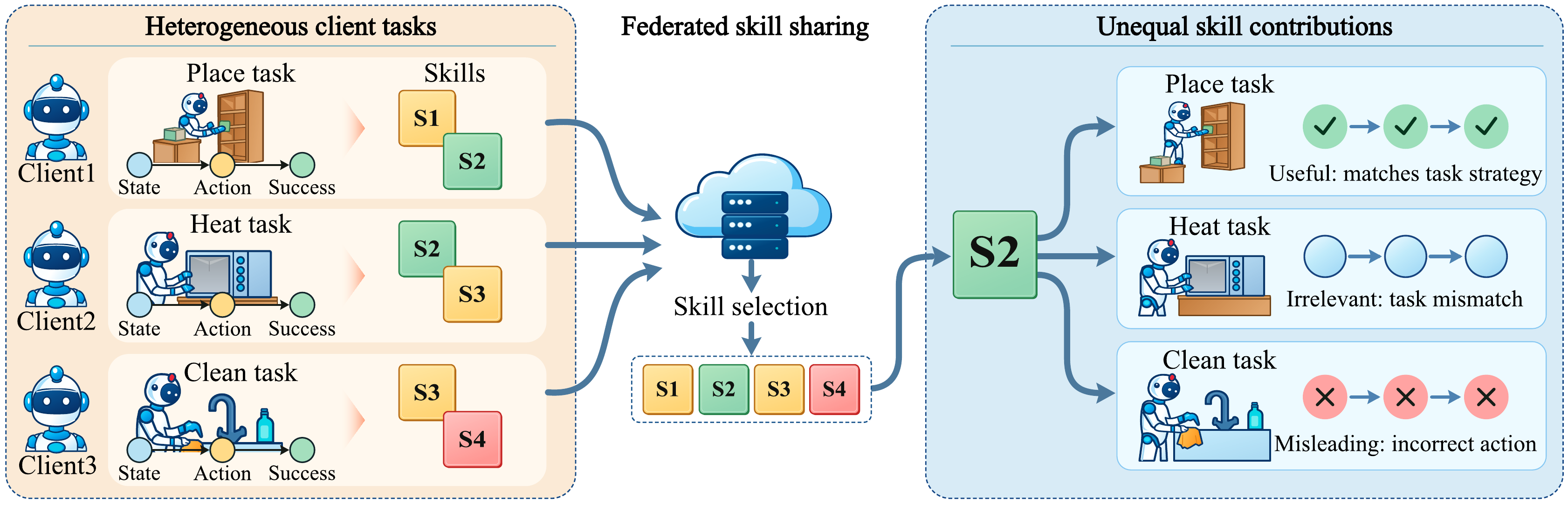}
    \caption{\rmfamily Illustration of unequal skill contributions under task heterogeneity.}
    \label{fig:FSL}
\end{figure}

To address this issue, federated skill learning (FSL) has emerged as a solution by distilling trajectories into compact skill representations and uploading them to a central server for sharing. For example, Lan et al.~\cite{lan2026fedskillgit} proposed FedSkillGit, which employs version control to maintain a global skill library and update shared skills based on the execution strategies derived from clients’ local tasks.
However, due to task heterogeneity, similar skills may contribute differently to task execution across clients, as illustrated in Fig.~\ref{fig:FSL}. For some clients, a shared skill can provide useful execution strategies, whereas for others, the same skill may be irrelevant to their tasks or even mislead the agent into taking incorrect actions~\cite{wang2025awm}. Ignoring such differences in skill utility across clients may result in suboptimal skill sharing, making it difficult for clients to prioritize the skills most relevant to their tasks and ultimately degrading agent performance.

To combat the above challenge, in this paper, we propose a
\underline{p}ersonalized \underline{fed}erated \underline{skill} learning framework,
named P-FedSkill, to provide each client with a personalized skill set under heterogeneous task distributions.
P-FedSkill consists of three key components: i) local skill extraction (LSE),
ii) personalized skill selection (PSS), and iii) task-aware skill routing (TASR).
Specifically, LSE groups local trajectories by task type and extracts skill patches (i.e., task applicability conditions and execution strategies) along with task execution outcomes. 
PSS then uses these outcomes to evaluate each skill patch's performance on client tasks and incorporates its task applicability condition to construct a personalized skill set for each client.
Finally, TASR adaptively selects a task-relevant subset of skill patches based on their applicability conditions to guide agent execution.
The main contributions of this paper are summarized as follows.

\begin{itemize}
    \item We propose P-FedSkill, a personalized federated skill learning framework that enables personalized skill sharing under task heterogeneity across clients.

    \item We design LSE to extract skill patches from each client's local trajectories, providing diverse solutions to guide agent execution.

    \item We design PSS to construct a personalized skill set for each client, and devise TASR to adaptively select task-relevant skill patches from the personalized skill set to guide agent execution.

    \item We establish a finite-sample suboptimality bound that provides a theoretical guarantee for personalized skill selection in P-FedSkill under heterogeneous client tasks, and extensive experiments further demonstrate its effectiveness.
\end{itemize}

The rest of the paper is organized as follows.
Sec.~\ref{sec:related_work} reviews the related work.
Sec.~\ref{sec:system_design} introduces the system design of P-FedSkill.
Sec.~\ref{sec:experiments} describes the experimental setup.
Sec.~\ref{sec:results_analysis} presents the results and analysis.
Finally, Sec.~\ref{sec:conclusion} concludes the paper.

\section{Related work}
\label{sec:related_work}
\subsection{Agent Skill Learning}

Skill learning has emerged as a promising method for improving agent performance by extracting reusable skills from interaction trajectories~\cite{fu2024language,zhou2025proposer,ding2026skillgen}.
Research efforts on agent skill learning predominantly focus on two major directions: experience abstraction and structured skill construction.
For experience-based skill learning, \cite{zhao2024expel} proposed ExpeL, which extracts reusable insights from successful and failed interaction trajectories and retrieves relevant experiences for unseen tasks. For structured skill construction, \cite{klissarov2025maestromotif} proposed MaestroMotif, which represents skills with initiation, execution, and termination components and uses LLM-generated code and reinforcement learning to construct and combine reusable skills.
Beyond centralized skill learning, recent studies have explored federated skill learning to enable cross-client skill sharing without exposing private interaction trajectories.
\cite{jiang2023fedskill} proposed FedSkill to enable cross-client skill sharing by clustering parameterized vectors based on similarity and returning matched cluster centroids to each client without sharing raw trajectories.

\begin{figure*}[t]
    \centering
    \includegraphics[width=0.85\textwidth]{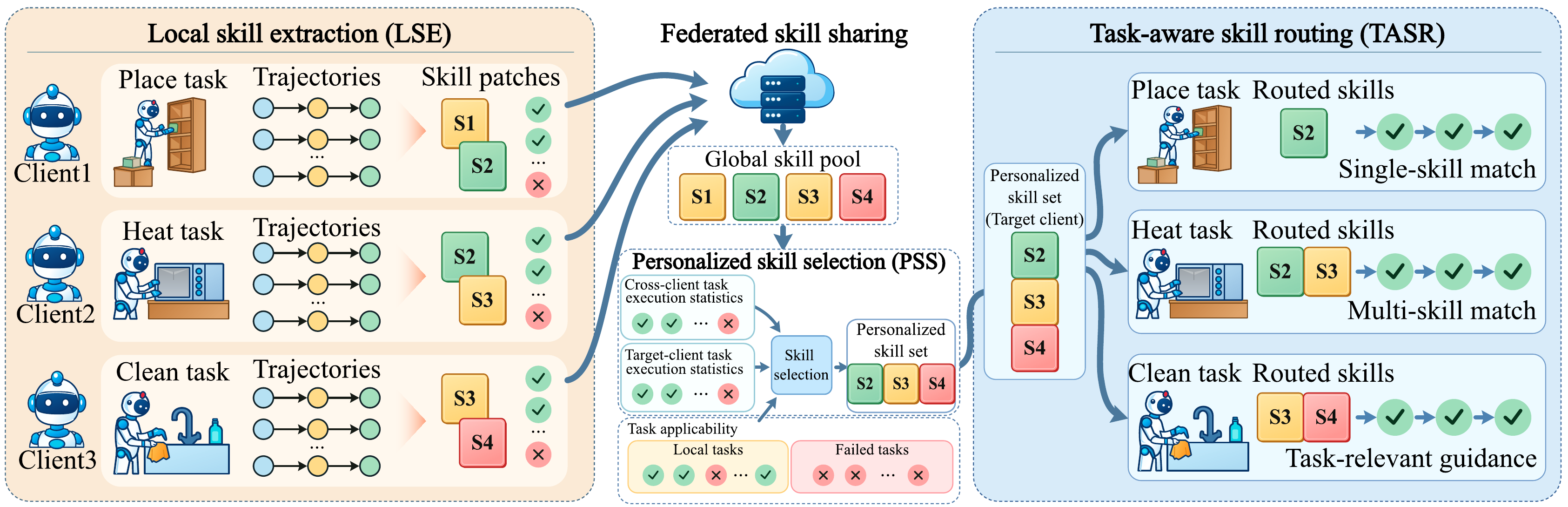}
    \caption{\centering Illustration of the P-FedSkill framework.}
    \label{fig:PFSL}
\end{figure*}

\subsection{Personalized Federated Learning under Heterogeneity}
Personalized federated learning (PFL) has been extensively studied to address client heterogeneity in federated learning~\cite{shen2025hotpluggable,zheng2025fedcalm,huang2025federated,liu2025understanding,yu2025understanding,wei2025fedwba}.
For example, \cite{shen2025hotpluggable} proposed HPFL, which trains personalized plug-in modules on a shared backbone and dynamically selects suitable modules for target data at inference.
Meanwhile, \cite{liu2025understanding} analyzed the generalization of PFL under non-convex objectives using uniform stability and decomposed the generalization error into aggregation and fine-tuning errors.
While these studies have advanced personalized federated learning under client heterogeneity, they primarily address model-level personalization or theoretical analysis rather than reusable skill sharing across agents.
As a result, personalized federated skill learning tailored to heterogeneous client tasks remains largely unexplored.

\section{System Design}
\label{sec:system_design}
\subsection{Overview}

In this section, we present the architecture and workflow of the P-FedSkill framework. As illustrated in Fig.~\ref{fig:PFSL}, P-FedSkill consists of three key modules: local skill extraction (LSE), personalized skill selection (PSS), and task-aware skill routing (TASR).
The workflow of P-FedSkill comprises the following four stages:
i) Each client interacts with its local environment to generate interaction trajectories;
ii) LSE then extracts skill patches and task execution statistics from these trajectories and uploads them to the central server;
iii) The uploaded skill patches are then processed by PSS to evaluate their performance on client tasks and construct a personalized skill set for each client.
The resulting personalized skill sets are then sent back to the corresponding clients;
iv) Each client receives its personalized skill set and employs TASR to select the relevant skill patches to guide agent execution.
The complete procedure of P-FedSkill is summarized in \textbf{Algorithm~\ref{alg:pfedskill}}.

\RestyleAlgo{ruled}
\LinesNumbered
\begin{algorithm}[h]
\caption{P-FedSkill framework.}
\footnotesize
\label{alg:pfedskill}
\setstretch{1}
\SetKwInOut{Input}{Require}
\SetKwInOut{Output}{Output}
\SetKwProg{Fn}{}{:}{}
\SetKwFunction{lse}{Local skill extraction}
\SetKwFunction{pss}{Personalized skill selection}
\SetKwFunction{tasr}{Task-aware skill routing}

\Input{$\{\mathcal{T}_k\}$, $\{d_k^t\}$, $\phi$, $\rho$,
$\epsilon_0$, $\eta$, $K$, $\delta$, $M_l$, $M_h$.}
\Output{$\{S_k\}$, $\{S_k^t\}$.}

\Fn{\lse}{
    \ForEach{client $k$}{
        \For{$j=1,\dots,z$}{
            $\mathcal{T}_{k,j}\gets
            \{\tau_{k,i}\in\mathcal{T}_k\mid \ell_{k,i}=j\}$\;
            \lIf{$\mathcal{T}_{k,j}\neq\emptyset$}{
                Extract skill patches $s=(c_s,\pi_s)$ from $\mathcal{T}_{k,j}$}
        }

        % Merge duplicate skill patches with identical $c_s$ and $\pi_s$\;

        \ForEach{skill patch $s$}{
            $\mathcal{E}_{k,s}\gets
            \{(i,u)\mid\text{subgoal }(i,u)\text{ uses }\pi_s,\ 
            y_{k,i,u}\text{ is observed}\}$,
            $n_{k,s}^{+}\gets\sum_{(i,u)\in\mathcal{E}_{k,s}}y_{k,i,u}$,
            $n_{k,s}^{-}\gets\sum_{(i,u)\in\mathcal{E}_{k,s}}(1-y_{k,i,u})$\;
        }

        Upload $\{(s,n_{k,s}^{+},n_{k,s}^{-})\}_s$ and
        $\{(\phi(d_{k,i}),y_{k,i})\}_{\tau_{k,i}\in\mathcal{T}_k}$
        to the server\;
    }
}

\Fn{\pss}{
    Consolidate uploaded skill patches into $\mathcal{S}$ and
    set $q_s\gets\phi(c_s)$ for all $s\in\mathcal{S}$\;

    \ForEach{client $k$}{
        $t_k\gets
        \operatorname{MeanPool}
        \big(\{\phi(d_{k,i})\}_{\tau_{k,i}\in\mathcal{T}_k}\big)$\;

        \lIf{$\exists\,i:\tau_{k,i}\in\mathcal{T}_k,\ y_{k,i}=0$}{
            $f_k\gets
            \operatorname{MeanPool}
            \big(\{\phi(d_{k,i})\mid\tau_{k,i}\in\mathcal{T}_k,\ y_{k,i}=0\}\big)$}

        \ForEach{$s\in\mathcal{S}$}{
            \lIf{$\mathcal{E}_{k,s}=\emptyset$}{
                $n_{k,s}^{+}\gets0$, $n_{k,s}^{-}\gets0$}

            $\mathcal{K}_{s}^{-k}\gets
            \{j\neq k\mid \mathcal{E}_{j,s}\neq\emptyset\}$,
            $\hat{p}_{-k,s}\gets
            \frac{
                \sum_{j\in\mathcal{K}_{s}^{-k}}n_{j,s}^{+}+\epsilon_0
            }{
                \sum_{j\in\mathcal{K}_{s}^{-k}}
                (n_{j,s}^{+}+n_{j,s}^{-})+2\epsilon_0
            }$\;

            $\alpha_{k,s}\gets
            \eta\hat{p}_{-k,s}+n_{k,s}^{+}$,
            $\beta_{k,s}\gets
            \eta(1-\hat{p}_{-k,s})+n_{k,s}^{-}$\;

            $\mu_{k,s}\gets
            \frac{\alpha_{k,s}}{\alpha_{k,s}+\beta_{k,s}}$,
            $\sigma_{k,s}\gets
            \sqrt{
                \frac{\alpha_{k,s}\beta_{k,s}}
                {(\alpha_{k,s}+\beta_{k,s})^2
                (\alpha_{k,s}+\beta_{k,s}+1)}
            }$\;

            $\tilde{p}_{k,s}\gets
            \max\{0,\mu_{k,s}-\sigma_{k,s}\}$\;

            \eIf{$\exists\,i:\tau_{k,i}\in\mathcal{T}_k,\ y_{k,i}=0$}{
                $a_{k,s}\gets
                \rho(q_s,t_k)+\rho(q_s,f_k)$\;
            }{
                $a_{k,s}\gets\rho(q_s,t_k)$\;
            }

            $v_{k,s}\gets\tilde{p}_{k,s}+a_{k,s}$\;
        }

        $S_k\gets
        \operatorname{TopK}_{s\in\mathcal{S}}\{v_{k,s}\}$\;
    }
}

\Fn{\tasr}{
    \ForEach{client $k$ and current task $t$}{
        $x_k^t\gets\phi(d_k^t)$\;

        \ForEach{$s\in S_k$}{
            $R_{k,s}^t\gets
            \rho(q_s,x_k^t)+\rho(q_s,t_k)$,
            $r_{k,s}^t\gets\rho(q_s,x_k^t)$\;
        }

        $\{s_{k,(i)}^t\}_{i=1}^{|S_k|}
        \gets
        \operatorname{sort}_{s\in S_k}^{\downarrow}
        \{R_{k,s}^t\}$\;

        \eIf{$|S_k|=1$}{
            $m_k^t\gets1$\;
        }{
            $r_{k,(1)}^t,r_{k,(2)}^t
            \gets
            \operatorname{Top2}_{s\in S_k}\{r_{k,s}^t\}$,
            $\Delta_k^t\gets
            r_{k,(1)}^t-r_{k,(2)}^t$\;

            $m_k^t\gets
            \begin{cases}
            \min\{M_h,|S_k|\}, & \Delta_k^t<\delta,\\
            M_l, & \Delta_k^t\ge\delta.
            \end{cases}$\;
        }

        $S_k^t\gets
        \{s_{k,(i)}^t\}_{i=1}^{m_k^t}$\;
    }
}
\end{algorithm}

\subsection{Local Skill Extraction}
\label{sec:lse}
As discussed in Sec.~\ref{sec:intro}, the narrow range of task types covered by local trajectories restricts the generalization of learned skills, thereby limiting agent performance. Given the diverse task solutions across clients, a straightforward solution is to leverage trajectories from multiple clients to learn more generalizable skills. However, raw trajectories often contain privacy-sensitive information, and directly collecting and sharing them may pose serious privacy risks~\cite{vekaria2025big}, making this solution impractical for real-world deployment.

\begin{figure}[t]
    \centering
    \subfloat[Steps.\label{fig:lse_steps}]{
    \begin{minipage}[b]{0.45\linewidth}
        \centering
        \hspace*{-1.5em}
        \includegraphics[width=\linewidth]{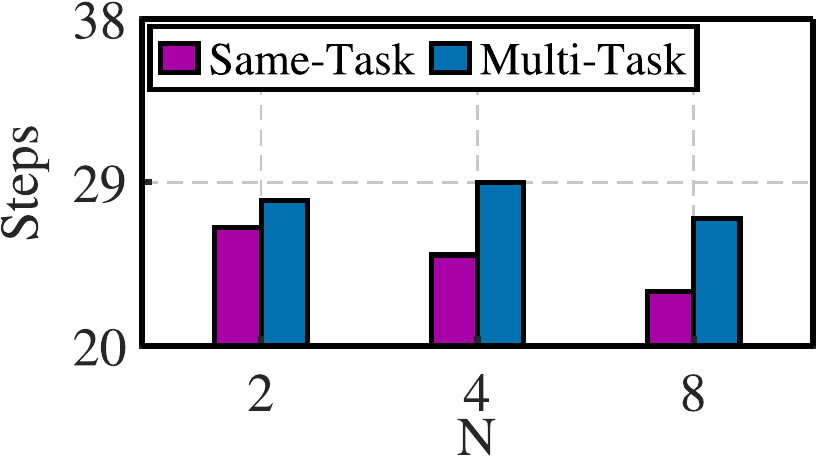}
    \end{minipage}}
    \hspace{0.03\linewidth}
    \subfloat[Accuracy.\label{fig:lse_accuracy}]{
    \begin{minipage}[b]{0.45\linewidth}
        \centering
        \hspace*{-1.5em}
        \includegraphics[width=\linewidth]{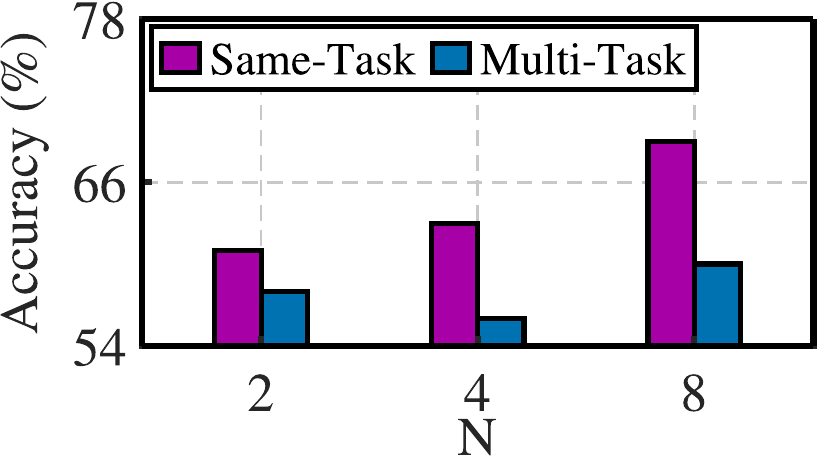}
    \end{minipage}}
    \caption{The impact of trajectory grouping on task completion steps and accuracy.}
    \label{fig:lse_preliminary}
\end{figure}

To address this challenge, we propose local skill extraction (LSE), which transforms trajectories into compact, reusable skill patches.
Since local trajectories may span different task types, extracting skills from trajectories across multiple task types may introduce irrelevant or even conflicting execution strategies, thereby degrading agent performance.
To investigate this insight, we conduct experiments on ALFWorld by varying the number of trajectories $N$ for skill extraction.
As shown in Fig.~\ref{fig:lse_steps}, skills extracted from same-task trajectories lead to fewer steps for task completion as the number of trajectories increases.
Fig.~\ref{fig:lse_accuracy} further shows that such skills achieve higher accuracy than those extracted from multi-task trajectories.
Based on these observations, LSE uses the agent to group each client's local trajectories based on their task descriptions and execution processes. Specifically, for client $k$, the agent assigns a category label $\ell_{k,i}$ to each trajectory $\tau_{k,i}$ and partitions its trajectory set $\mathcal{T}_k$ into $z$ disjoint groups:
\begin{equation}
\mathcal{T}_{k,j}
=
\left\{
\tau_{k,i}\in\mathcal{T}_k
\mid
\ell_{k,i}=j
\right\},
\end{equation}
where $\ell_{k,i} \in \{1,\ldots,z\}$ denotes the agent-assigned category of trajectory $\tau_{k,i}$, and
$\mathcal{T}_{k,j}$ denotes the trajectory group corresponding to the $j$-th category of client $k$.

After grouping, each trajectory group contains trajectories with similar task descriptions or execution patterns.
Since raw trajectories contain extensive interaction details, directly using them to guide the agent may introduce task-irrelevant information, thereby reducing task execution efficiency.
To mitigate this issue, LSE extracts skill patches from each trajectory group, with each skill patch comprising a task applicability condition and an execution strategy.
Specifically, within each trajectory group $\mathcal{T}_{k,j}$, LSE uses the agent to extract the task applicability condition $c_s$ from the grouped trajectories and task descriptions, derive the execution strategy $\pi_s$ from the tool calls and action sequences shared across the trajectories while filtering trajectory-specific operations, and record the observable outcomes of subgoals executing the corresponding strategy, yielding the skill patch $s$ as
\begin{equation}
s
=
\left(
c_s,
\pi_s
\right).
\end{equation}

Since different trajectories within the same group may adopt the same task solution, the above skill extraction process can produce redundant skill patches with identical task applicability conditions and execution strategies.
To avoid redundant skill storage and sharing, LSE merges these duplicate skill patches and aggregates their corresponding subgoal execution outcomes.
For a given skill patch $s$ on client $k$, the numbers of successful and failed subgoal instances, denoted by $n_{k,s}^{+}$ and $n_{k,s}^{-}$, are given by
\begin{equation}
n_{k,s}^{+}
=
\sum_{(i,u)\in\mathcal{E}_{k,s}} y_{k,i,u},
\qquad
n_{k,s}^{-}
=
\sum_{(i,u)\in\mathcal{E}_{k,s}}
\left(1-y_{k,i,u}\right),
\end{equation}
where $\mathcal{E}_{k,s}$ denotes the set of subgoal indices $(i,u)$ corresponding to executions of strategy $\pi_s$ with observable outcomes in client $k$'s source trajectories, and $y_{k,i,u}\in\{0,1\}$ denotes the outcome of the $u$-th subgoal in trajectory $\tau_{k,i}$, with 1 and 0 indicating subgoal success and failure, respectively.
Finally, each client uses the frozen text encoder $\phi(\cdot)$~\cite{chen2024m3} to encode the task description $d_{k,i}$ associated with $\tau_{k,i}$, and uploads the resulting representations to the server together with the corresponding skill patches and subgoal execution statistics.

\subsection{Personalized Skill Selection}
\label{sec:pss}
\begin{figure}[t]
    \centering
    \subfloat[Client performance.\label{fig:psa_motivation1}]{
    \begin{minipage}[b]{0.45\linewidth}
        \centering
        \hspace*{-1.5em}
        \includegraphics[width=\linewidth]{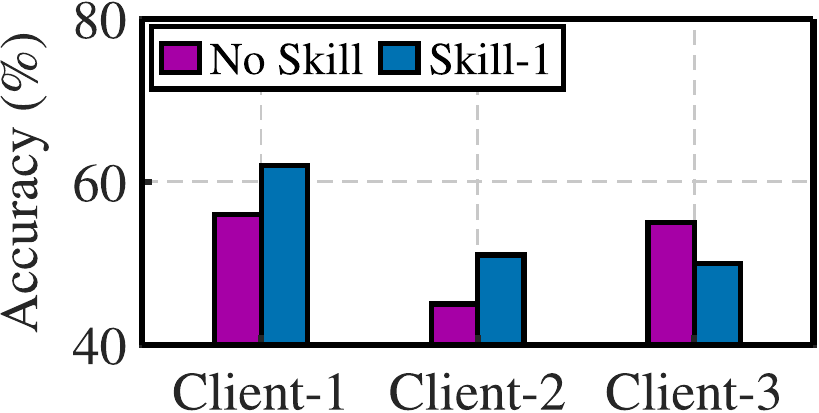}
    \end{minipage}}
    \hspace{0.03\linewidth}
    \subfloat[Best skill patch.\label{fig:psa_motivation2}]{
    \begin{minipage}[b]{0.45\linewidth}
        \centering
        \hspace*{-1.5em}
        \includegraphics[width=\linewidth]{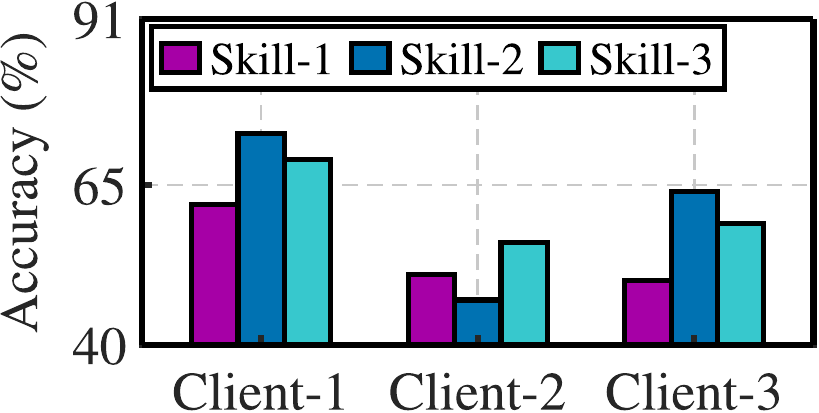}
    \end{minipage}}

    \caption{The impact of skill patches on agent performance across different clients.}

    \label{fig:psa_preliminary}
\end{figure}

As described in Sec.~\ref{sec:lse}, LSE transforms local trajectories into reusable skill patches, enabling diverse task solutions to be shared across clients. However, due to the heterogeneity of client tasks, the same skill patch may contribute differently to agent performance across clients. To investigate this impact, we conduct experiments on ALFWorld by providing the same skill patch to different clients. Fig.~\ref{fig:psa_motivation1} shows that the same skill patch does not improve agent performance on every client.
Fig.~\ref{fig:psa_motivation2} further shows that the best-performing skill patch varies across clients. Therefore, it is imperative to perform personalized skill selection for each client.

Motivated by these observations, we propose personalized skill selection (PSS) to evaluate the performance of each skill patch on client tasks based on its associated subgoal execution outcomes and incorporate its task applicability condition to construct a personalized skill set for each client. PSS first collects skill patches uploaded by different clients and uses the agent to consolidate semantically equivalent skill patches to construct a global skill pool $\mathcal{S}$, while retaining their subgoal execution statistics. Since the number of subgoal instances associated with a skill patch may vary across clients, estimates based on limited local subgoal instances are sensitive to individual outcomes. The subgoal execution statistics from other clients provide additional information for the same skill patch, thereby improving the reliability of its evaluation~\cite{thapa2025harnessing}. Accordingly, PSS evaluates each skill patch using the cross-client performance score $\hat{p}_{-k,s}$, defined as
\begin{equation}
\hat{p}_{-k,s}
=
\frac{
\sum_{j\in\mathcal{K}_{s}^{-k}} n_{j,s}^{+}
+\varepsilon_0
}{
\sum_{j\in\mathcal{K}_{s}^{-k}}
\left(
n_{j,s}^{+}+n_{j,s}^{-}
\right)
+2\varepsilon_0
},
\label{eq:psa_cross_client}
\end{equation}
where $\mathcal{K}_{s}^{-k}$ denotes the set of source clients for skill patch $s$, excluding target client $k$, and $\epsilon_0>0$ is a smoothing constant.

Building on $\hat{p}_{-k,s}$, PSS further uses subgoal execution outcomes of the target client to evaluate the contribution of skill patches to local tasks. Specifically, PSS first constructs a Beta prior for each skill patch based on $\hat{p}_{-k,s}$:
\begin{equation}
\theta_{k,s}
\sim
\operatorname{Beta}
\left(
\eta\hat{p}_{-k,s},
\eta(1-\hat{p}_{-k,s})
\right),
\end{equation}
where $\theta_{k,s}$ denotes the success probability of subgoals corresponding to skill patch $s$ on client $k$, and $\eta$ denotes the prior strength.
Since the Beta distribution is conjugate to the binomial likelihood, the Beta prior based on \(\hat{p}_{-k,s}\) can be updated with the target client's local subgoal execution outcomes for adaptive personalized skill evaluation. For clients with fewer local subgoal instances, the evaluation relies more on the cross-client prior to reduce the influence of individual subgoal outcomes, while relying more on local subgoal execution outcomes as more subgoal instances become available. PSS then uses the subgoal execution outcomes
$D_{k,s}=\{y_{k,i,u}\}_{(i,u)\in\mathcal{E}_{k,s}}$
of the target client to update the prior of $\theta_{k,s}$, yielding the posterior distribution
\begin{equation}
\theta_{k,s}\mid D_{k,s}
\sim
\operatorname{Beta}
\left(
\eta\hat{p}_{-k,s}+n_{k,s}^{+},
\eta(1-\hat{p}_{-k,s})+n_{k,s}^{-}
\right).
\end{equation}

Based on the posterior distribution, the posterior mean $\mu_{k,s}$ estimates the success probability of subgoals corresponding to skill patch $s$ on the target client, while the posterior standard deviation $\sigma_{k,s}$ quantifies its uncertainty. However, due to the limited number of local subgoal instances, the posterior estimate may exhibit high uncertainty, thereby reducing the reliability of the skill evaluation. Therefore, PSS penalizes $\mu_{k,s}$ by $\sigma_{k,s}$ to obtain the performance score $\tilde{p}_{k,s}$:
\begin{equation}
\tilde{p}_{k,s}
=
\max\{0,\mu_{k,s}-\sigma_{k,s}\}.
\label{eq:personalized_performance}
\end{equation}

Although Beta-based skill evaluation can mitigate the impact of limited local subgoal instances, a higher $\tilde{p}_{k,s}$ does not necessarily indicate greater potential to improve agent performance, as the corresponding skill patch may not match the client's local tasks or help address its failed tasks. To address this limitation, PSS further evaluates the applicability of each skill patch to local and failed tasks as
\begin{equation}
a_{k,s}
=
\rho(q_s,t_k)
+
\rho(q_s,f_k),
\end{equation}
where $a_{k,s}$ denotes the applicability score of skill patch $s$ for client $k$; $q_s$ denotes the feature representation of the task applicability condition $c_s$ obtained using the frozen text encoder $\phi(\cdot)$; $t_k$ and $f_k$ denote the aggregated feature representations of all local and failed local task descriptions of client $k$, respectively, obtained via the same encoder and mean pooling; and $\rho(\cdot,\cdot)$ denotes the normalized cosine similarity, linearly scaled to $[0,1]$. For clients without failed tasks, the failed-task relevance term is set to zero.
PSS then combines the performance score and applicability score for personalized skill selection:
\begin{equation}
v_{k,s}
=
\tilde{p}_{k,s}
+
a_{k,s},
\end{equation}
where $v_{k,s}$ denotes the personalized score of skill patch $s$ for client $k$.
Finally, PSS ranks the skill patches collected by the server according to $v_{k,s}$ and selects the top-$K$ patches to construct the personalized skill set $\mathcal{S}_k$ for client $k$.

\subsection{Task-Aware Skill Routing}

\label{sec:tasr}
Although the personalized skill set constructed by PSS provides each client with task-related skill patches, under task heterogeneity, a client may encounter multiple task types that require different execution strategies. Therefore, it is necessary to adaptively select relevant skill patches from the personalized skill set to guide the client agent for task execution.  

To this end, we propose task-aware skill routing (TASR), which adaptively selects appropriate skill patches from the personalized skill set based on their relevance to the current task. Specifically, for the task $t$ with task description $d_k^t$, TASR first obtains its feature representation $x_k^t$ via $\phi(\cdot)$. It then measures the relevance of the task applicability condition ${q}_s$ to ${x}_k^t$ and ${t}_k$, and combines the two relevance terms to compute the task-aware routing score of skill patch $s$:
\begin{equation}
R_{k,s}^{t}
=
\rho\!\left(q_s,x_k^{t}\right)
+
\rho\!\left(q_s,t_k\right).
\end{equation}

A larger $R_{k,s}^{t}$ indicates a higher routing priority. Based on the task-aware routing scores, the skill patches in the personalized skill set $\mathcal{S}_k$ are sorted in descending order of routing priority, yielding the ordered skill sequence $\{s_{k,(1)}^{t},s_{k,(2)}^{t},\ldots,s_{k,(|\mathcal{S}_k|)}^{t}\}$, given by
\begin{equation}
\left\{
s_{k,(1)}^{t},
s_{k,(2)}^{t},
\ldots,
s_{k,(|\mathcal{S}_k|)}^{t}
\right\}
=
\underset{s\in\mathcal{S}_k}{\operatorname{sort}_{d}}
\left\{
R_{k,s}^{t}
\right\},
\end{equation}
where $s_{k,(i)}^{t}$ denotes the skill patch with the $i$-th highest task-aware routing score, and $\operatorname{sort}_{d}(\cdot)$ returns the corresponding skill patches in descending order of their scores.  

Due to task heterogeneity, client tasks may require different numbers of relevant skill patches for effective execution. Using a fixed number of skill patches may introduce redundant strategies under clear skill matching or omit potentially useful strategies under ambiguous skill matching.
Therefore, it is necessary to select skill patches for different client tasks adaptively.
The task-aware routing score reflects skill-task matching, while the gap between the top relevance scores provides a signal for adaptive skill routing. A substantially higher task relevance score indicates a clear match with the current task, suggesting that only a small number of highly relevant skills are needed for effective execution. In contrast, similar task relevance scores across multiple skill patches indicate ambiguous skill matching with the current task, suggesting that more candidate skills should be retained to avoid discarding potentially useful execution strategies. Therefore, for a personalized skill set containing only one skill patch, TASR directly routes the sole patch. Otherwise, we use the gap $\Delta_k^t$ between the highest and second-highest task relevance scores as the routing criterion, given by
\begin{equation}
\Delta_k^{t}
=
r_{k,(1)}^{t}
-
r_{k,(2)}^{t},
\end{equation}
where $r_{k,s}^{t}=\rho(q_s,x_k^{t})$ denotes the relevance of skill patch $s$ to the current task, and $r_{k,(1)}^{t}$ and $r_{k,(2)}^{t}$ denote the highest and second-highest task relevance scores, respectively. Based on $\Delta_k^{t}$, TASR adaptively determines the number of routed skill patches as
\begin{equation}
m_k^t =
\begin{cases}
\min\{M_h,|\mathcal{S}_k|\}, & \Delta_k^t < \delta,\\
M_l, & \Delta_k^t \geq \delta.
\end{cases}
\end{equation}
where $\delta$ is the predefined relevance-gap threshold, and $1 \leq M_l < M_h$ denote the smaller and larger numbers of routed skill patches, respectively. TASR then selects the top-$m_k^{t}$ skill patches from the ordered skill sequence to construct $\mathcal{S}_k^{t}=\{s_{k,(1)}^{t},s_{k,(2)}^{t},\ldots,s_{k,(m_k^{t})}^{t}\}$ to guide agent execution.

\subsection{Theoretical Analysis}

\label{sec:theory}

We analyze the personalized skill selection of PSS under heterogeneous client tasks. For client $k$ and skill patch $s$, let $n_{k,s}=n_{k,s}^{+}+n_{k,s}^{-}$. The conditional oracle scores are $\tilde{p}_{k,s}^{*}=\theta_{k,s}$ and $v_{k,s}^{*}=\tilde{p}_{k,s}^{*}+a_{k,s}$.
For $M$ skill patches with $1\leq K\leq M$, $\mathcal{S}_k$ and $\mathcal{S}_k^{*}$ contain the top-$K$ patches according to $v_{k,s}$ and $v_{k,s}^{*}$, respectively, with $\epsilon_k=\max_s\epsilon_{k,s}$.

% ==========================================
% Assumptions
% ==========================================

\begin{assumption}
\label{assump:bernoulli}
The subgoal execution outcome satisfies
$y_{k,i,u}\mid\theta_{k,s}\sim
\operatorname{Bernoulli}(\theta_{k,s})$
for $(i,u)\in\mathcal{E}_{k,s}$.
\end{assumption}

\begin{assumption}
\label{assump:independence}
The subgoal execution outcomes
$\{y_{k,i,u}\}_{(i,u)\in\mathcal{E}_{k,s}}$
are conditionally independent given $\theta_{k,s}$.
\end{assumption}

\begin{assumption}
\label{assump:prior}
The prior strength satisfies $0<\eta<\infty$.
\end{assumption}

% ==========================================
% Lemma 1 and its proof
% ==========================================

\begin{lemma}
\label{lem:posterior_mean}
Under Assumptions~\ref{assump:bernoulli}--\ref{assump:prior},
for any $\xi\in(0,1)$, with probability at least $1-\xi$,
\begin{equation}
\left|\mu_{k,s}-\theta_{k,s}\right|
\leq
\frac{
\eta\left|\hat{p}_{-k,s}-\theta_{k,s}\right|
+
\sqrt{\frac{n_{k,s}}{2}\log\frac{2}{\xi}}
}{
\eta+n_{k,s}
}.
\label{eq:mu_bound}
\end{equation}
\end{lemma}

\begin{IEEEproof}
From Eq.~(6), the posterior mean is
\begin{equation}
\mu_{k,s}
=
\frac{
\eta\hat{p}_{-k,s}+n_{k,s}^{+}
}{
\eta+n_{k,s}
}.
\label{eq:mu_mean}
\end{equation}
Hence,
\begin{equation}
\mu_{k,s}-\theta_{k,s}
=
\frac{
\eta(\hat{p}_{-k,s}-\theta_{k,s})
+
(n_{k,s}^{+}-n_{k,s}\theta_{k,s})
}{
\eta+n_{k,s}
}.
\label{eq:mu_decomp}
\end{equation}
By the triangle inequality,
\begin{equation}
\left|
\mu_{k,s}-\theta_{k,s}
\right|
\leq
\frac{
\eta\left|\hat{p}_{-k,s}-\theta_{k,s}\right|
}{
\eta+n_{k,s}
}
+
\frac{
\left|
n_{k,s}^{+}-n_{k,s}\theta_{k,s}
\right|
}{
\eta+n_{k,s}
}.
\label{eq:mu_triangle}
\end{equation}

For $n_{k,s}>0$, by Eq.~(3),
\begin{equation}
n_{k,s}^{+}
=
\sum_{(i,u)\in\mathcal{E}_{k,s}}y_{k,i,u}.
\label{eq:success_sum}
\end{equation}
Under Assumptions~\ref{assump:bernoulli} and
\ref{assump:independence}, Hoeffding's inequality gives
\begin{equation}
\Pr\left(
\left|
n_{k,s}^{+}
-
n_{k,s}\theta_{k,s}
\right|
\geq
\sqrt{
\frac{n_{k,s}}{2}
\log\frac{2}{\xi}
}
\right)
\leq
\xi.
\label{eq:hoeffding}
\end{equation}

Thus, with probability at least $1-\xi$,
\begin{equation}
\left|
n_{k,s}^{+}
-
n_{k,s}\theta_{k,s}
\right|
\leq
\sqrt{
\frac{n_{k,s}}{2}
\log\frac{2}{\xi}
}.
\label{eq:local_bound}
\end{equation}

Substituting Eq.~(\ref{eq:local_bound}) into
Eq.~(\ref{eq:mu_triangle}) yields Eq.~(\ref{eq:mu_bound}).

For $n_{k,s}=0$, Eq.~(\ref{eq:mu_mean}) gives
$\mu_{k,s}=\hat{p}_{-k,s}$, and
Eq.~(\ref{eq:mu_bound}) holds directly.
\end{IEEEproof}

% ==========================================
% Lemma 2 and its proof
% ==========================================

\begin{lemma}
\label{lem:score_bound}
Under Assumptions~\ref{assump:bernoulli}--\ref{assump:prior},
let $M=|\mathcal{S}|$. For any $\xi\in(0,1)$,
with probability at least $1-\xi$,
\begin{equation}
\begin{aligned}
\left|v_{k,s}-v_{k,s}^{*}\right|
&\leq \epsilon_{k,s}
=
\frac{
\eta\left|\hat{p}_{-k,s}-\theta_{k,s}\right|
+
\sqrt{\frac{n_{k,s}}{2}\log\frac{2M}{\xi}}
}{
\eta+n_{k,s}
}\\
&\quad+
\frac{1}{2\sqrt{\eta+n_{k,s}+1}},
\qquad
\forall s\in\mathcal{S}.
\end{aligned}
\label{eq:v_bound}
\end{equation}
\end{lemma}

\begin{IEEEproof}
For the Beta posterior in Eq.~(6), let
\begin{equation}
\alpha_{k,s}
=
\eta\hat{p}_{-k,s}+n_{k,s}^{+},
\qquad
\beta_{k,s}
=
\eta(1-\hat{p}_{-k,s})+n_{k,s}^{-}.
\label{eq:beta_parameters}
\end{equation}
Then,
\begin{equation}
\alpha_{k,s}+\beta_{k,s}
=
\eta+n_{k,s}.
\label{eq:beta_sum}
\end{equation}

The posterior variance is
\begin{equation}
\sigma_{k,s}^{2}
=
\frac{
\alpha_{k,s}\beta_{k,s}
}{
(\alpha_{k,s}+\beta_{k,s})^{2}
(\alpha_{k,s}+\beta_{k,s}+1)
}.
\label{eq:beta_variance}
\end{equation}

Since
\begin{equation}
\alpha_{k,s}\beta_{k,s}
\leq
\frac{
(\alpha_{k,s}+\beta_{k,s})^{2}
}{4},
\end{equation}
we obtain
\begin{equation}
\sigma_{k,s}
\leq
\frac{
1
}{
2\sqrt{\eta+n_{k,s}+1}
}.
\label{eq:sigma_bound}
\end{equation}

Since $\theta_{k,s}\in[0,1]$, we have
$\tilde{p}_{k,s}^{*}=\theta_{k,s}
=\max\{0,\theta_{k,s}\}$.
Using the $1$-Lipschitz property of
$\max\{0,x\}$ in Eq.~(7) gives
\begin{align}
\left|
\tilde{p}_{k,s}-\tilde{p}_{k,s}^{*}
\right|
&\leq
\left|
\mu_{k,s}
-
\sigma_{k,s}
-
\theta_{k,s}
\right|
\nonumber\\
&\leq
\left|
\mu_{k,s}-\theta_{k,s}
\right|
+\sigma_{k,s}.
\label{eq:p_bound}
\end{align}

Applying Lemma~\ref{lem:posterior_mean} with confidence
parameter $\xi/M$ and taking a union bound over the
$M$ skill patches yields, with probability at least
$1-\xi$,
\begin{equation}
\left|
\mu_{k,s}-\theta_{k,s}
\right|
\leq
\frac{
\eta\left|\hat{p}_{-k,s}-\theta_{k,s}\right|
+
\sqrt{
\frac{n_{k,s}}{2}
\log\frac{2M}{\xi}
}
}{
\eta+n_{k,s}
},
\qquad
\forall s\in\mathcal{S}.
\label{eq:uniform_mu_bound}
\end{equation}

Substituting Eq.~(\ref{eq:sigma_bound}) and
Eq.~(\ref{eq:uniform_mu_bound}) into
Eq.~(\ref{eq:p_bound}) gives
\begin{equation}
\left|
\tilde{p}_{k,s}-\tilde{p}_{k,s}^{*}
\right|
\leq
\epsilon_{k,s},
\qquad
\forall s\in\mathcal{S}.
\label{eq:p_final_bound}
\end{equation}

Since $v_{k,s}$ and $v_{k,s}^{*}$ share the same
applicability score $a_{k,s}$,
\begin{equation}
\left|
v_{k,s}-v_{k,s}^{*}
\right|
=
\left|
\tilde{p}_{k,s}-\tilde{p}_{k,s}^{*}
\right|
\leq
\epsilon_{k,s},
\qquad
\forall s\in\mathcal{S}.
\label{eq:v_final_bound}
\end{equation}
\end{IEEEproof}

% ==========================================
% Theorem and its proof
% ==========================================

\begin{theorem}
\label{thm:pss}
Under Assumptions~\ref{assump:bernoulli}--\ref{assump:prior},
define
\begin{equation}
\epsilon_k
\triangleq
\max_{s\in\mathcal{S}}\epsilon_{k,s}.
\label{eq:epsilon_k}
\end{equation}
For any $\xi\in(0,1)$, with probability at least
$1-\xi$, the personalized skill selection of PSS
satisfies the following finite-sample suboptimality bound:
\begin{equation}
0
\leq
\sum_{s\in\mathcal{S}_k^{*}}v_{k,s}^{*}
-
\sum_{s\in\mathcal{S}_k}v_{k,s}^{*}
\leq
2K\epsilon_k.
\label{eq:selection_bound}
\end{equation}
\end{theorem}

\begin{IEEEproof}
By Lemma~\ref{lem:score_bound} and the definition
of $\epsilon_k$, with probability at least $1-\xi$,
\begin{equation}
\left|
v_{k,s}-v_{k,s}^{*}
\right|
\leq
\epsilon_k,
\qquad
\forall s\in\mathcal{S}.
\label{eq:uniform_v_bound}
\end{equation}

Since $\mathcal{S}_k$ contains the top-$K$ skill
patches according to $v_{k,s}$,
\begin{equation}
\sum_{s\in\mathcal{S}_k^{*}}
v_{k,s}
\leq
\sum_{s\in\mathcal{S}_k}
v_{k,s}.
\label{eq:topk_property}
\end{equation}

Using Eq.~(\ref{eq:uniform_v_bound}) and
Eq.~(\ref{eq:topk_property}), we obtain
\begin{align}
\sum_{s\in\mathcal{S}_k^{*}}
v_{k,s}^{*}
&\leq
\sum_{s\in\mathcal{S}_k^{*}}
v_{k,s}
+
K\epsilon_k
\nonumber\\
&\leq
\sum_{s\in\mathcal{S}_k}
v_{k,s}
+
K\epsilon_k
\nonumber\\
&\leq
\sum_{s\in\mathcal{S}_k}
v_{k,s}^{*}
+
2K\epsilon_k.
\label{eq:selection_upper}
\end{align}

Since $\mathcal{S}_k^{*}$ contains the top-$K$
skill patches according to $v_{k,s}^{*}$,
\begin{equation}
\sum_{s\in\mathcal{S}_k^{*}}
v_{k,s}^{*}
\geq
\sum_{s\in\mathcal{S}_k}
v_{k,s}^{*}.
\label{eq:selection_lower}
\end{equation}

Combining Eq.~(\ref{eq:selection_upper}) and
Eq.~(\ref{eq:selection_lower}), we obtain
\begin{equation}
0
\leq
\sum_{s\in\mathcal{S}_k^{*}}
v_{k,s}^{*}
-
\sum_{s\in\mathcal{S}_k}
v_{k,s}^{*}
\leq
2K\epsilon_k.
\label{eq:selection_final}
\end{equation}
\end{IEEEproof}

\section{Experimental Setup}
\label{sec:experiments}

\textbf{Datasets and Model.}
We adopt the  ALFWorld~\cite{shridhar2021alfworld} and $\tau^2$-Bench~\cite{barres2026tau2} datasets to evaluate the performance of P-FedSkill.
Experiments are conducted under independent and identically distributed (IID) and non-IID settings.
For the IID setting, tasks are randomly shuffled and evenly distributed across clients.
For the non-IID setting, tasks are partitioned using a Dirichlet distribution with concentration parameter $\alpha$.
We use Qwen3.6-35B-A3B~\cite{qwen36_35b_a3b} as the shared backbone agent for all methods and keep its model architecture and inference configuration fixed across experiments to ensure a fair comparison.

\textbf{Baselines.}
We evaluate P-FedSkill on $\tau^2$-Bench under both IID and non-IID settings by comparing it with the backbone agent operating without learned skills (FSL-B), FederatedSkill~\cite{yang2026federatedskill}, FedSkillGit~\cite{lan2026fedskillgit}, and AWM~\cite{wang2025awm}.

\textbf{Hyper-parameters.}
In our experiments, we set the number of clients to 5.
For the non-IID setting, the Dirichlet concentration parameter $\alpha$ is set to 0.1.
For PSS, $\epsilon_0$ and $\eta$ are set to 1 and 8, respectively.
For a fair comparison, the maximum candidate skill budget is set to $K=\lceil 0.6|\mathcal{S}|\rceil$ for all skill-based methods.

\section{Results and Analysis}
\label{sec:results_analysis}
\subsection{Superiority of P-FedSkill}
As demonstrated in Tab.~\ref{tab:tau_bench_accuracy}, P-FedSkill outperforms all baselines in all three domains of $\tau^2$-Bench under both IID and non-IID settings.
Specifically, P-FedSkill achieves accuracies of 66.67\% (80.00\%), 76.47\% (70.59\%), and 76.47\% (79.41\%) in the airline, retail, and telecom domains, respectively, under the IID (non-IID) setting.
This performance advantage stems from personalized skill set construction and task-aware skill routing, which better align shared skills with heterogeneous client tasks.
In contrast, FederatedSkill merges skill patches extracted from local trajectories into client-specific skill libraries. Since tasks within the same client may require different skills, patches learned from previous tasks may introduce steps that are irrelevant to the current task, thereby degrading agent performance.

\begin{table}[t]
\centering
\small
\caption{Accuracy (\%) of P-FedSkill and baselines on $\tau^2$-Bench under IID and non-IID settings.}
\label{tab:tau_bench_accuracy}
\setlength{\tabcolsep}{4pt}
\renewcommand{\arraystretch}{1.05}
\begin{adjustbox}{max width=\linewidth}
\begin{tabular}{l ccc ccc}
\toprule
& \multicolumn{3}{c}{IID}
& \multicolumn{3}{c}{Non-IID} \\
\cmidrule(lr){2-4}
\cmidrule(lr){5-7}
Method & Airline & Retail & Telecom & Airline & Retail & Telecom \\
\midrule
\textbf{P-FedSkill}
& \textbf{66.67} & \textbf{76.47} & \textbf{76.47}
& \textbf{80.00} & \textbf{70.59} & \textbf{79.41} \\
FederatedSkill
& 46.67 & 73.53 & 64.71 & 73.33 & 64.71 & 67.65 \\
FedSkillGit
& 46.67 & 67.65 & 58.82 & 66.67 & 61.76 & 64.71 \\
FSL-B
& 53.33 & 64.71 & 58.82 & 53.33 & 64.71 & 58.82 \\
AWM
& 60.00 & 64.71 & 58.82 & 60.00 & 64.71 & 58.82 \\
\bottomrule
\end{tabular}
\end{adjustbox}
\end{table}

\section{Conclusion}
\label{sec:conclusion}
In this paper, we have proposed a personalized federated skill learning framework, named P-FedSkill, aimed at facilitating personalized skill sharing under task heterogeneity across clients.
P-FedSkill comprises three key components: LSE, PSS, and TASR.
LSE extracts skill patches and task execution outcomes from trajectories.
PSS then uses these outcomes and task applicability conditions to construct a personalized skill set for each client, while TASR adaptively selects task-relevant skill patches to guide agent execution.
We further establish a finite-sample suboptimality bound, providing a theoretical guarantee for personalized skill selection in P-FedSkill. Extensive experiments also demonstrate the effectiveness of P-FedSkill.

\bibliographystyle{IEEEtran}
\bibliography{reference}

\end{document}